\documentclass[10pt,twocolumn,letterpaper]{article}
\usepackage{cvpr} 
\usepackage[accsupp]{axessibility}  

\usepackage{graphicx}
\usepackage{amsmath}
\usepackage{amssymb}
\usepackage{booktabs}
\usepackage{multirow}
\usepackage{enumitem}
\usepackage{array}
\usepackage{makecell}
\usepackage{microtype}
\usepackage{caption}
\usepackage{subcaption}
\usepackage{circledsteps}
\usepackage[numbers,sort,compress]{natbib}
\usepackage[pagebackref,breaklinks,colorlinks,bookmarks=false]{hyperref}

\usepackage[capitalize]{cleveref}
\crefname{section}{Sec.}{Secs.}
\Crefname{section}{Section}{Sections}
\Crefname{table}{Table}{Tables}
\crefname{table}{Tab.}{Tabs.}
\usepackage{booktabs,caption}
\def\cvprPaperID{6224} 
\def\confName{CVPR}
\def\confYear{2023}
\def\Approach{ISBNet}

\begin{document}
\def\mA{\mathcal{A}}
\def\mB{\mathcal{B}}
\def\mC{\mathcal{C}}
\def\mD{\mathcal{D}}
\def\mE{\mathcal{E}}
\def\mF{\mathcal{F}}
\def\mG{\mathcal{G}}
\def\mH{\mathcal{H}}
\def\mI{\mathcal{I}}
\def\mJ{\mathcal{J}}
\def\mK{\mathcal{K}}
\def\mL{\mathcal{L}}
\def\mM{\mathcal{M}}
\def\mN{\mathcal{N}}
\def\mO{\mathcal{O}}
\def\mP{\mathcal{P}}
\def\mQ{\mathcal{Q}}
\def\mR{\mathcal{R}}
\def\mS{\mathcal{S}}
\def\mT{\mathcal{T}}
\def\mU{\mathcal{U}}
\def\mV{\mathcal{V}}
\def\mW{\mathcal{W}}
\def\mX{\mathcal{X}}
\def\mY{\mathcal{Y}}
\def\mZ{\mathcal{Z}} 

\def\bbN{\mathbb{N}} 
\def\bbR{\mathbb{R}} 
\def\bbP{\mathbb{P}} 
\def\bbQ{\mathbb{Q}} 
\def\bbE{\mathbb{E}}

\def\1n{\mathbf{1}_n}
\def\0{\mathbf{0}}
\def\1{\mathbf{1}}

\def\A{{\bf A}}
\def\B{{\bf B}}
\def\C{{\bf C}}
\def\D{{\bf D}}
\def\E{{\bf E}}
\def\F{{\bf F}}
\def\G{{\bf G}}
\def\H{{\bf H}}
\def\I{{\bf I}}
\def\J{{\bf J}}
\def\K{{\bf K}}
\def\L{{\bf L}}
\def\M{{\bf M}}
\def\N{{\bf N}}
\def\O{{\bf O}}
\def\P{{\bf P}}
\def\Q{{\bf Q}}
\def\R{{\bf R}}
\def\S{{\bf S}}
\def\T{{\bf T}}
\def\U{{\bf U}}
\def\V{{\bf V}}
\def\W{{\bf W}}
\def\X{{\bf X}}
\def\Y{{\bf Y}}
\def\Z{{\bf Z}}

\def\a{{\bf a}}
\def\b{{\bf b}}
\def\c{{\bf c}}
\def\d{{\bf d}}
\def\e{{\bf e}}
\def\f{{\bf f}}
\def\g{{\bf g}}
\def\h{{\bf h}}
\def\i{{\bf i}}
\def\j{{\bf j}}
\def\k{{\bf k}}
\def\l{{\bf l}}
\def\m{{\bf m}}
\def\n{{\bf n}}
\def\o{{\bf o}}
\def\p{{\bf p}}
\def\q{{\bf q}}
\def\r{{\bf r}}
\def\s{{\bf s}}
\def\t{{\bf t}}
\def\u{{\bf u}}
\def\v{{\bf v}}
\def\w{{\bf w}}
\def\x{{\bf x}}
\def\y{{\bf y}}
\def\z{{\bf z}}

\def\balpha{\mbox{\boldmath{$\alpha$}}}
\def\bbeta{\mbox{\boldmath{$\beta$}}}
\def\bdelta{\mbox{\boldmath{$\delta$}}}
\def\bgamma{\mbox{\boldmath{$\gamma$}}}
\def\blambda{\mbox{\boldmath{$\lambda$}}}
\def\bsigma{\mbox{\boldmath{$\sigma$}}}
\def\btheta{\mbox{\boldmath{$\theta$}}}
\def\bomega{\mbox{\boldmath{$\omega$}}}
\def\bxi{\mbox{\boldmath{$\xi$}}}
\def\bnu{\mbox{\boldmath{$\nu$}}}                                  
\def\bphi{\mbox{\boldmath{$\phi$}}}
\def\bmu{\mbox{\boldmath{$\mu$}}}

\def\bDelta{\mbox{\boldmath{$\Delta$}}}
\def\bOmega{\mbox{\boldmath{$\Omega$}}}
\def\bPhi{\mbox{\boldmath{$\Phi$}}}
\def\bLambda{\mbox{\boldmath{$\Lambda$}}}
\def\bSigma{\mbox{\boldmath{$\Sigma$}}}
\def\bGamma{\mbox{\boldmath{$\Gamma$}}}
                                  
\newcommand{\myprob}[1]{\mathop{\mathbb{P}}_{#1}}

\newcommand{\myexp}[1]{\mathop{\mathbb{E}}_{#1}}

\newcommand{\mydelta}[1]{1_{#1}}

\newcommand{\myminimum}[1]{\mathop{\textrm{minimum}}_{#1}}
\newcommand{\mymaximum}[1]{\mathop{\textrm{maximum}}_{#1}}    
\newcommand{\mymin}[1]{\mathop{\textrm{minimize}}_{#1}}
\newcommand{\mymax}[1]{\mathop{\textrm{maximize}}_{#1}}
\newcommand{\mymins}[1]{\mathop{\textrm{min.}}_{#1}}
\newcommand{\mymaxs}[1]{\mathop{\textrm{max.}}_{#1}}  
\newcommand{\myargmin}[1]{\mathop{\textrm{argmin}}_{#1}} 
\newcommand{\myargmax}[1]{\mathop{\textrm{argmax}}_{#1}} 
\newcommand{\myst}{\textrm{s.t. }}

\newcommand{\denselist}{\itemsep -1pt}
\newcommand{\sparselist}{\itemsep 1pt}

\definecolor{pink}{rgb}{0.9,0.5,0.5}
\definecolor{purple}{rgb}{0.5, 0.4, 0.8}   
\definecolor{gray}{rgb}{0.3, 0.3, 0.3}
\definecolor{mygreen}{rgb}{0.2, 0.6, 0.2}

\newcommand{\cyan}[1]{\textcolor{cyan}{#1}}
\newcommand{\red}[1]{\textcolor{red}{#1}}  
\newcommand{\blue}[1]{\textcolor{blue}{#1}}
\newcommand{\magenta}[1]{\textcolor{magenta}{#1}}
\newcommand{\pink}[1]{\textcolor{pink}{#1}}
\newcommand{\green}[1]{\textcolor{green}{#1}} 
\newcommand{\gray}[1]{\textcolor{gray}{#1}}    
\newcommand{\mygreen}[1]{\textcolor{mygreen}{#1}}    
\newcommand{\purple}[1]{\textcolor{purple}{#1}}       

\definecolor{greena}{rgb}{0.4, 0.5, 0.1}
\newcommand{\greena}[1]{\textcolor{greena}{#1}}

\definecolor{bluea}{rgb}{0, 0.4, 0.6}
\newcommand{\bluea}[1]{\textcolor{bluea}{#1}}
\definecolor{reda}{rgb}{0.6, 0.2, 0.1}
\newcommand{\reda}[1]{\textcolor{reda}{#1}}

\def\changemargin#1#2{\list{}{\rightmargin#2\leftmargin#1}\item[]}
\let\endchangemargin=\endlist
                                               
\newcommand{\cm}[1]{}

\newcommand{\mhoai}[1]{{\color{magenta}\textbf{[MH: #1]}}}

\newcommand{\mtodo}[1]{{\color{red}$\blacksquare$\textbf{[TODO: #1]}}}
\newcommand{\myheading}[1]{\vspace{1ex}\noindent \textbf{#1}}
\newcommand{\htimesw}[2]{\mbox{$#1$$\times$$#2$}}


\newif\ifshowsolution
\showsolutiontrue

\ifshowsolution  
\newcommand{\Comment}[1]{\paragraph{\bf $\bigstar $ COMMENT:} {\sf #1} \bigskip}
\newcommand{\Solution}[2]{\paragraph{\bf $\bigstar $ SOLUTION:} {\sf #2} }
\newcommand{\Mistake}[2]{\paragraph{\bf $\blacksquare$ COMMON MISTAKE #1:} {\sf #2} \bigskip}
\else
\newcommand{\Solution}[2]{\vspace{#1}}
\fi

\newcommand{\truefalse}{
\begin{enumerate}
	\item True
	\item False
\end{enumerate}
}

\newcommand{\yesno}{
\begin{enumerate}
	\item Yes
	\item No
\end{enumerate}
}

\newcommand{\Sref}[1]{Sec.~\ref{#1}}
\newcommand{\Eref}[1]{Eq.~(\ref{#1})}
\newcommand{\Fref}[1]{Fig.~\ref{#1}}
\newcommand{\Tref}[1]{Table~\ref{#1}}

\definecolor{mydarkblue}{rgb}{0,0.08,1}
\definecolor{mydarkgreen}{rgb}{0.02,0.6,0.02}
\definecolor{myred}{rgb}{1.0,0.0,0.0}
\newcommand{\khoi}[1]{\textcolor{mydarkblue}{[Khoi: #1]}}
\newcommand{\tuan}[1]{\textcolor{mydarkgreen}{[Tuan: #1]}}
\newcommand{\son}[1]{\textcolor{myred}{[Son: #1]}}

\title{ISBNet: a 3D Point Cloud Instance Segmentation Network \\ with Instance-aware Sampling and Box-aware Dynamic Convolution}

\author{Tuan Duc Ngo \qquad Binh-Son Hua \qquad Khoi Nguyen\\
VinAI Research, Hanoi, Vietnam\\
{\tt\small \{v.tuannd42, v.sonhb, v.khoindm\}@vinai.io}
}

\maketitle
\thispagestyle{plain}
\pagestyle{plain}
\begin{abstract}
    Existing 3D instance segmentation methods are predominated by the bottom-up design -- manually fine-tuned algorithm to group points into clusters followed by a refinement network. However, by relying on the quality of the clusters, these methods generate susceptible results when (1) nearby objects with the same semantic class are packed together, or (2) large objects with loosely connected regions. To address these limitations, we introduce \Approach, a novel cluster-free method that represents instances as kernels and decodes instance masks via dynamic convolution. To efficiently generate high-recall and discriminative kernels, we propose a simple strategy named Instance-aware Farthest Point Sampling to sample candidates and leverage the local aggregation layer inspired by PointNet++ to encode candidate features. Moreover, we show that predicting and leveraging the 3D axis-aligned bounding boxes in the dynamic convolution further boosts performance. Our method set new state-of-the-art results on ScanNetV2 (55.9), S3DIS (60.8), and STPLS3D (49.2) in terms of AP and retains fast inference time (237ms per scene on ScanNetV2). The source code and trained models are available at \url{https://github.com/VinAIResearch/ISBNet}.
    
\end{abstract}

\vspace{-12pt}
\section{Introduction}
\label{sec:intro}

\begin{figure}[t]
  \centering
  \includegraphics[width=0.95\linewidth]{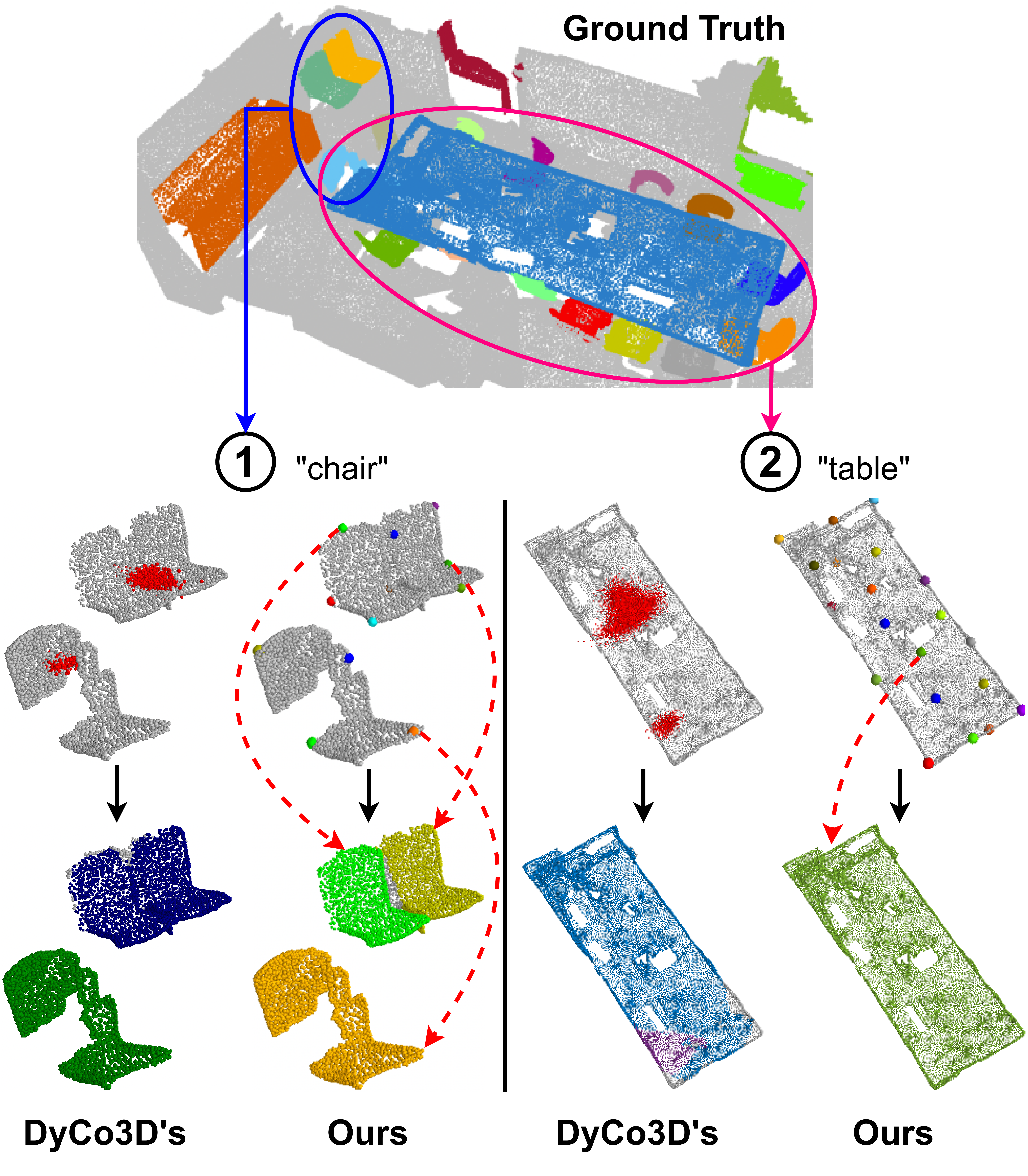}
  \vspace{-10pt}
   \caption{
   In DyCo3D \cite{he2021dyco3d}, kernel prediction quality is greatly affected by the centroid-based clustering algorithm which has two issues:
   \textbf{\Circled[]{1}} mis-grouping nearby instances and \textbf{\Circled[]{2}} over-segment a large object into multiple fragments. Our method addresses these issues by \emph{instance-aware point sampling}, achieving far better results. 
   Each sample point aggregates information from its local context to generate a kernel for predicting its own object mask, and the final instances will be filtered and selected by an NMS. 
   }
   \label{fig:demo}
   \vspace{-16pt}
\end{figure}

3D instance segmentation (3DIS) is a core problem of deep learning in the 3D domain. Given a 3D scene represented by a point cloud, we seek to assign each point with a semantic class and a unique instance label.
3DIS is an important 3D perception task and has a wide range of applications in autonomous driving, augmented reality, and robot navigation where point cloud data can be leveraged to complement the information provided by 2D images. Compared to 2D image instance segmentation (2DIS), 3DIS is arguably harder due to much higher variations in appearance and spatial extent along with unequal distribution of point cloud, i.e., dense near object surface and sparse elsewhere. Thus, it is not trivial to apply 2DIS methods to 3DIS. 

\begin{figure*}[t]
  \centering
  \includegraphics[width=0.95\linewidth]{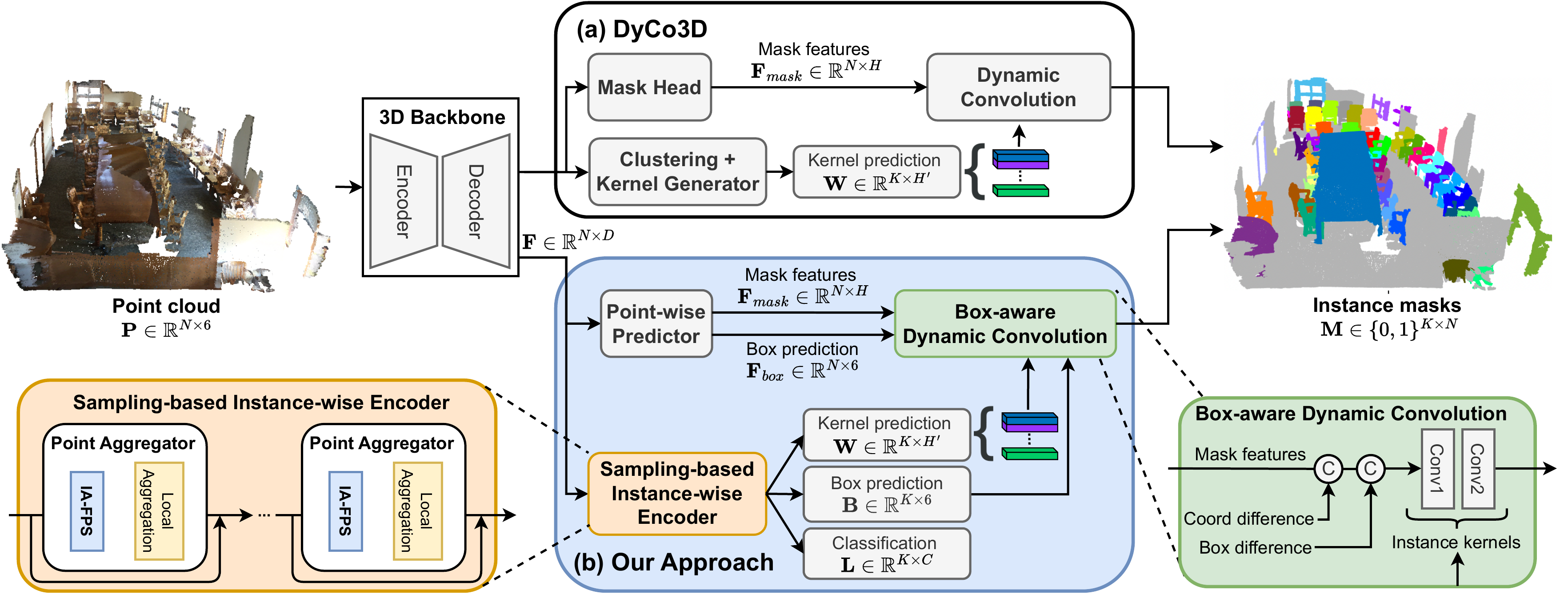}
   \vspace{-4pt}
   \caption{Overall architectures of DyCo3D \cite{he2021dyco3d} (block (a)) and our approach (block (b)) for 3DIS. Given a point cloud, a 3D backbone is employed to extract per-point features. 
   For DyCo3D, it first groups points into clusters based on the predicted object centroid from each point to generate a kernel for each cluster. In the meantime, the mask head transforms the per-point features into mask features for dynamic convolution.
   For our approach \Approach, we replace the clustering algorithm with a novel sampling-based instance-wise encoder to obtain faster and more robust kernel, box, and class predictions. Furthermore, a point-wise predictor replaces the mask head of DyCo3D to output the mask and box features for a new box-aware dynamic convolution to produce more accurate instance masks.
   }
   \label{fig:architecture}
   \vspace{-16pt}
\end{figure*}

A typical approach for 3DIS, DyCo3D \cite{he2021dyco3d}, adopts dynamic convolution \cite{wang2020solov2, tian2020conditional} to predict instance masks. Specifically, points are clustered, voxelized, and passed through a 3D Unet to generate instance kernels for dynamic convolution with the feature of all points in the scene. This approach is illustrated in Fig.~\ref{fig:architecture}~(a). However, this approach has several limitations.
First, the clustering algorithm heavily relies on the centroid-offset prediction whose quality deteriorates significantly when:
(1) objects are densely packed so that two objects can be mistakenly grouped together as one object, or (2) large objects whose parts are loosely connected resulting in different objects when clustered. These two scenarios are visualized in Fig.~\ref{fig:demo}. 
Second, the points' feature mostly encodes object appearance which is not distinct enough for separating different instances, especially between objects having the same semantic class.


To address the limitations of DyCo3D \cite{he2021dyco3d}, we propose \Approach, a cluster-free framework for 3DIS with \textbf{I}nstance-aware Farthest Point \textbf{S}ampling and \textbf{B}ox-aware Dynamic Convolution.
First, we revisit the Farthest Point Sampling (FPS) \cite{eldar1997farthest} and the clustering method in \cite{he2021dyco3d,chen2021hierarchical,vu2022softgroup} and find that these algorithms generate considerably low instance recall. As a result, many objects are omitted in the subsequent stage, leading to poor performance. Motivated by this, we propose our Instance-aware Farthest Point Sampling (IA-FPS), which aims to sample query candidates in a 3D scene with high instance recall. We then introduce our Point Aggregator, incorporating the IA-FPS with a local aggregation layer to encode semantic features, shapes, and sizes of instances into instance-wise features. 

Additionally, the 3D bounding box of the object is an existing supervision but has not yet been explored in the 3D instance segmentation task. Therefore, we add an auxiliary branch to our model to jointly predict the axis-aligned bounding box and the binary mask of each instance. The ground-truth axis-aligned bounding box is deduced from the existing instance mask label. Unlike Mask-DINO \cite{li2022mask} and CondInst \cite{tian2020conditional}, where the auxiliary bounding box prediction is just used as a regularization of the learning process, we leverage it as an extra geometric cue in the dynamic convolution, thus further boosting the performance of the instance segmentation task.

To evaluate the performance of our approach, we conduct extensive experiments on three challenging datasets: ScanNetV2 \cite{dai2017scannet}, S3DIS \cite{armeni2017joint}, and STPLS3D \cite{chen2022stpls3d}. \Approach~not only achieves the highest accuracy among these three datasets, surpassing the strongest method by +2.7/2.4/3.0 on ScanNetV2, S3DIS, and STPLS3D, but also demonstrates to be highly efficient, running at 237ms per scene on ScanNetV2.

In summary, the contributions of our work are as follows:

\begin{itemize}[noitemsep,topsep=0pt,leftmargin=*]
    \item We propose \Approach, 
    a cluster-free paradigm for 3DIS, that leverages Instance-aware Farthest Point Sampling and Point Aggregator to generate an instance feature set.
    \item We first introduce using the axis-aligned bounding box as an auxiliary supervision and propose the Box-aware Dynamic Convolution to decode instance binary masks.
    \item \Approach~achieves state-of-the-art performance on three different datasets: ScanNetV2, S3DIS, and STPLS3D without comprehensive modifications of the model architecture and hyper-parameter tuning for each dataset.
\end{itemize}

In the following, Sec.~\ref{sec:related_work} reviews prior work; Sec.~\ref{sec:approach} specifies our approach; and Sec.~\ref{sec:experiments} presents our implementation details and experimental results. Sec.~\ref{sec:conclusion} concludes with some remarks and discussions.

\section{Related Work}
\label{sec:related_work}

\myheading{2D image instance segmentation (2DIS)} concerns assigning each pixel in the image with one of the instance labels and semantic labels. Its approaches can be divided into three groups: proposal-based, proposal-free, and DETR-based approaches. For proposal-based methods \cite{he2017mask,Cai2018CascadeRD,kirillov2020pointrend}, an object detector, e.g., Faster-RCNN \cite{ren2015fasterrcnn} is leveraged to predict object bounding boxes to segment the foreground region inside detected boxes.
For proposal-free methods \cite{wang2020solo,wang2020solov2,tian2020conditional}, SOLO \cite{wang2020solo,wang2020solov2} and CondInst \cite{tian2020conditional} predict the instance kernels for the dynamic convolution with the feature maps to generate instance masks. For DETR-based methods \cite{guo2021sotr,Cheng2021PerPixelCI,Cheng2022MaskedattentionMT,li2022mask}, Mask2Former \cite{Cheng2022MaskedattentionMT} and Mask-DINO \cite{li2022mask} employ the transformer architecture with instance queries to obtain the segmentation for each instance. 
Compared to 3DIS, 2DIS is arguably easier due to the structured, grid-based, and dense properties of 2D images. Hence, it is not trivial to adapt a 2DIS method to 3DIS.

\myheading{3D point cloud instance segmentation (3DIS)} methods are interested in labeling each point in a 3D point cloud with a semantic class and a unique instance ID. They can be categorized into proposal-based, clustering-based, and dynamic convolution-based methods. Cluster

\textit{Proposal-based methods} \cite{hou20193d,yang2019learning,yi2019gspn} first detect 3D bounding boxes and then segment the foreground region inside each box to form instances. 3D-SIS \cite{hou20193d} adapts the Mask R-CNN architecture to 3D instance segmentation and jointly learns features from two modalities of RGB images and 3D point clouds. 
3D-BoNet \cite{yang2019learning} predicts a fixed number of 3D bounding boxes from a global feature vector summarizing the content of the scene and then segments the foreground points inside each box. 
A limitation of this approach is that the performance of instance masks heavily depends on the quality of 3D bounding boxes which is very unstable due to the huge variation and uneven distribution of 3D point cloud.
%
%

\textit{Clustering-based methods} \cite{wang2018sgpn,jiang2020pointgroup,chen2021hierarchical,he2021dyco3d,engelmann20203d,pham2019jsis3d,vu2022softgroup,zhao2022divide} learn latent embeddings that facilitate grouping points into instances. 
PointGroup \cite{jiang2020pointgroup} predicts the 3D offset from each point to its instance's centroid and obtains the clusters from two point clouds: original points and centroid-shifted points. 
HAIS \cite{chen2021hierarchical} proposes a hierarchical clustering method where a small cluster can be filtered out or absorbed by a larger cluster. 
SoftGroup \cite{vu2022softgroup} proposes a soft-grouping strategy in which each point can belong to multiple clusters with different semantic classes to alleviate the semantic prediction error. 
One of the limitations of the clustering-based approach is that the quality of the instance masks significantly depends on the quality of the clustering, i.e., the centroid prediction, which is greatly unreliable especially when testing objects considerably differ from training objects in spatial extent.

\textit{Dynamic convolution-based methods} \cite{he2021dyco3d,He2022PointInst3DS3,wu2022dknet} overcome the limitations of proposal-based and clustering-based methods by generating kernels and then using them to convolve with the point features to generate instance masks. 
DyCo3D \cite{he2021dyco3d} adopts the clustering algorithm in \cite{jiang2020pointgroup} to generate kernels for dynamic convolution. 
PointInst3D \cite{He2022PointInst3DS3} uses farthest-point sampling to replace the clustering in \cite{he2021dyco3d} in order to generate kernels.
DKNet \cite{wu2022dknet} introduces candidate mining and candidate aggregation to generate more discriminative instance kernels for dynamic convolution.

Our approach is a dynamic convolution-based method with two important improvements in kernel generation and dynamic convolution. Particularly, in the former, we propose a new instance encoder combining instance-aware farthest-point sampling with a point aggregation layer to generate kernels to replace clustering in DyCo3D \cite{he2021dyco3d}. In the latter, instead of only using the appearance feature for dynamic convolution, we additionally enhance that feature with a geometry cue namely bounding box prediction. 

\section{Our Approach}

\label{sec:approach}

\myheading{Problem statement:}
Given a 3D point cloud $\mathbf{P} \in \mathbb{R}^{N\times6}$ where $N$ is the number of points, and each point is represented by a 3D position and RGB color vector.
We aim to segment the point cloud into $K$ instances that are represented by a set of binary masks $\mathbf{M} \in \{0,1\}^{K \times N}$ and a set of semantic labels $\mathbf{L} \in \mathbb{R} ^{K \times C}$, where $C$ is the number of semantic categories. 

Our method consists of four main components: a 3D backbone, a point-wise predictor, a sampling-based instance-wise encoder, and a box-aware dynamic convolution. 
The 3D backbone takes a 3D point cloud as input to extract per-point features. 
Our backbone network extracts feature $f^{(i)} \in \mathbb{R}^{D}$ where $i=1,\ldots, N$ for each point of the input point cloud.
We follow previous methods \cite{chen2021hierarchical,vu2022softgroup,wu2022dknet} to adopt a U-Net with sparse convolutions \cite{graham20183d} as our backbone. 
The point-wise predictor takes per-point features $\mathbf{F} \in \mathbb{R}^{N \times D}$ from the backbone and transforms them into point-wise semantic predictions, axis-aligned bounding box predictions $\mathbf{F}_{box} \in \mathbb{R}^{N \times 6}$, and mask features $\mathbf{F}_{mask} \in \mathbb{R}^{N \times H}$ for box-aware dynamic convolution.
The sampling-based instance-wise encoder (Sec.~\ref{sec:inst_encoder}) processes point-wise features to generate instance kernels, instance class labels, and bounding box parameters. 
Finally, the box-aware dynamic convolution (Sec.~\ref{sec:box_dyco}) gets the instance kernels and the mask features with the complementary box prediction to generate the final binary mask for each instance.
An overview of our method is illustrated in Fig.~\ref{fig:architecture}.

\begin{table}[t]
\small
\setlength{\tabcolsep}{7pt}
\centering
\begin{tabular}{lcccc}
\toprule
\textbf{\# sampling points}         & 2048  & 512  & 256 & 128  \\ 
\midrule
FPS      & 99.3\% &  93.3\% & 85.4\% & 71.3\% \\ 
IA-FPS  & 100\% & 98.4\% &  94.5\% &  89.2\% \\
\midrule
Clustering & 75.5\% & 75.5\% & 75.5\%& 75.5\% \\
\bottomrule
\end{tabular}
\vspace{-4pt}
\caption{The recall of different sampling methods on ScanNetV2 validation set. FPS is the standard Farthest Point Sampling, IA-FPS is our proposed Instance-aware Farthest Point Sampling. Clustering is the algorithm used in \cite{he2021dyco3d,chen2021hierarchical,vu2022softgroup} and its recall does not depend on the number of sampling points.}
\label{tab:recall_rate}
\vspace{-16pt}
\end{table}

\subsection{Sampling-based Instance-wise Encoder}
\label{sec:inst_encoder}



Given per-point features $\mathbf{F} \in \mathbb{R}^{N \times D}$ output from the backbone, 
we aim to produce instance-wise features $\mathbf{E} \in \mathbb{R}^{K \times D}$ where $K \ll N$. 
The instance-wise feature $\mathbf{E}$ is then used to predict the instance classification scores $\mathbf{L} \in \mathbb{R}^{K \times C}$, instance boxes $\mathbf{B} \in \mathbb{R}^{K \times 6}$, and instance kernels $\mathbf{W} \in \mathbb{R}^{K \times H'}$ where $H'$ is decided by the sizes of the convolutional layers in dynamic convolution. 

Typically, one can employ Farthest Point Sampling (FPS) \cite{eldar1997farthest} to sample a set of $K$ candidates to generate instance kernels as in \cite{He2022PointInst3DS3}. FPS greedily samples points in 3D coordinates by choosing the next points farthest away from the previous sampled ones using the pairwise distance.
However, this sampling technique is inferior.
First, there are many points belonging to the background categories among the $K$ sampled candidates by FPS, wasting computational resources. 
Second, large objects dominate the number of sampled points hence no point is sampled from small objects. Third, the point-wise features cannot capture the local context to create instance kernels. 
We provide analysis in Tab.~\ref{tab:recall_rate} to validate this observation.
Particularly, we calculate the recall of the number of instances predicted by the kernels against the total ground truth instances. 
The recall value should be large as we expect good coverage of the clustered or sampled points on the ground-truth instances. 
However, as can be seen, previous methods have low recall which can be explained by that these methods do not consider instances for point clustering or sampling. 

To address this issue, we propose a novel sampling-based instance-wise encoder that takes instances into account in the point sampling step. 
Inspired by the Set Abstraction in PointNet++ \cite{qi2018pointnet}, we specify our instance encoder comprising a sequence of Point Aggregator (PA) blocks whose components are Instance-Aware FPS (IA-FPS) to sample candidate points covering as many foreground objects as possible and a local aggregation layer to capture the local context so as to enrich the candidate features individually. We visualize the PA in the orange block in Fig.~\ref{fig:architecture} and detail our sampling below.





\myheading{Instance-aware FPS.} Our sampling strategy is to sample foreground points to maximally cover all instances regardless of their sizes. To achieve this goal, we opt for an iterative sampling technique as follows. Specifically, candidates are sampled from a set of points that are neither background nor chosen by previous sampled candidates. 
We use the point-wise semantic prediction to estimate the probability for each point to be background $m^{(i)}_{(0)} \in [0, 1]$. 
We also use the instance masks generated by previous $k$-th candidate $m^{(i)}_{(k)} \in [0, 1]$. 
The FPS is leveraged to sample points from the set of points $\mathbf{P'} \subset \mathbf{P}$: 
\begin{equation}
    \vspace{-4pt}
    \label{eq:sampling}
    \mathbf{P'} = \left \{p^{(i)} \in \mathbf{P} \bigg| \min_{k={0 .. K'}} \left(1 - m^{(i)}_{(k)} \right) > \tau\right \},
\end{equation}
where $K'$ is the number of already chosen candidates and $\tau$, is the hyper-parameter threshold. 

Practically, in training, since the instance mask prediction is not good enough for guiding the instance sampling, $K$ candidates are sampled altogether at once from the predicted foreground mask $1 -m^{(i)}_{(0)} > \tau$.
On the other hand, in testing, we iteratively sample smaller chunks $\{ \kappa_{1},\ldots, \kappa_{T}\}$ one by one such that the subsequent chunks will be sampled from neither the background points nor points belonging to predicted masks of previous chunks. By doing so, the recall rate of IA-FPS improves a lot as shown in Tab.~\ref{tab:recall_rate}.


\myheading{Local Aggregation Layer.} For each candidate $k$, the local aggregation layer encodes and transforms the local context into its instance-wise features. 
Specifically, Ball-query is employed \cite{qi2018pointnet} to collect its $Q$ local neighbors as the local features $\mathbf{F}_{local}^{(k)} \in \mathbb{R}^{Q \times D}$. Also, the relative coordinates between the candidates $k$ and their neighbors $q$ are computed and normalized with the neighborhood radius $r$ to form the local coordinates, or $\mathbf{P}_{local}^{(k)} \in [-1, 1]^{Q \times 3}$.
%
%
%
Next, we use an MLP layer to transform the local features $\mathbf{F}_{local}^{(k)}$ and the local coordinates $\mathbf{P}_{local}^{(k)}$ into the instance-wise features $e^{(k)}$ of candidate $k$. We also add a residual connection with the original features $f^{(k)}$ to avoid gradient vanishing. Concretely, the instance-wise feature can be computed as:
\begin{equation}
    \vspace{-4pt}
    e^{(k)} = f^{(k)} + \max_q \left ( \text{MLP}\left ( \left[\mathbf{F}_{local}^{(k)}; \mathbf{P}_{local}^{(k)} \right]\right ) \right ),
\end{equation}
%
where $\left[\cdot ; \cdot\right]$ denotes the concatenation operations. From $\mathbf{E}$, a linear layer is used to predict instance classification scores $\mathbf{L}$, instance boxes $\mathbf{B}$, and instance kernels $\mathbf{W}$.

It is worth noting that, to obtain the instance-wise features $\mathbf{E}$, instead of using a single PA block, we propose a progressive way, by sequentially applying multiple PA blocks. In this way, the subsequent block will sample from the smaller number of points sampled by the previous block. Doing so has the same effect as stacking multiple convolutional layers in 2D images in order to increase the receptive field. 
%



\subsection{Box-aware Dynamic Convolution}
\label{sec:box_dyco}





In the dynamic convolution of \cite{he2021dyco3d,He2022PointInst3DS3,wu2022dknet}, for each candidate $k$, the relative position of all points w.r.t $k$, $\mathbf{F}_{pos}^{(k)} \in \mathbb{R}^{N \times 3}$ and the point-wise mask features $\mathbf{F}_{mask} \in \mathbb{R}^{N \times H}$
are concatenated and convolved with instance kernels $w^{(k)}$ to obtain the instance binary mask $
    \widehat{m}^{(k)} = \text{Sigmoid}\left(\text{Conv} \left( \left[\mathbf{F}_{mask}; \mathbf{F}_{pos}^{(k)}\right]; w^{(k)} \right)\right),
$
where Conv is implemented as several convolutional layers. However, we would argue that only using the mask features and positions is sub-optimal. For example, points near the boundary of adjacent objects whose class is the same are indistinguishable from each other when only using the mask features and positions in 3D.
%

On the other hand, 3D bounding box delineates the shape and size of an object, which provides an important geometric cue for the prediction of object masks in instance segmentation. 
Our method uses bounding box predictions as an auxiliary task that regularizes instance segmentation training. 
Particularly, for each point, we propose to regress the axis-aligned bounding box deduced from the object mask. 
The predicted boxes $\mathbf{F}_{box} \in \mathbb{R}^{N \times 6} $ are then used to condition the mask feature to generate kernels for the box-aware dynamic convolution (see the green block in Fig.~\ref{fig:architecture}).
Each bounding box is parameterized by a 6D vector $f_{box}^{(i)}=(x_1,y_1,z_1,x_2,y_2,z_2)$ that represents the minimum and maximum bound of the point coordinates of an instance. It is worth noting that we choose to use axis-aligned bounding boxes because the ground-truth boxes are basically available for free as they can be easily constructed from ground-truth instance annotations. 

Therefore, we propose to use the predicted boxes as an additional geometric cue in dynamic convolution, giving the name of our proposed box-aware dynamic convolution. Intuitively, two points will belong to the same object if their predicted boxes are similar. 
%
Our final instance mask $\widehat{m}^{(k)} \in [0,1]^{1 \times N}$ of the $k$-th candidate is obtained as:
\begin{equation}
    \widehat{m}^{(k)} = \text{Sigmoid}\left(\text{Conv} \left( \left[\mathbf{F}_{mask}; \mathbf{F}_{pos}^{(k)};\mathbf{F}_{geo}^{(k)}\right]; w^{(k)} \right)\right).
\end{equation}
The geometric feature $\mathbf{F}_{geo}^{(k)} \in \mathbb{R}^{N \times 6}$ can be calculated from the absolute difference of the bounding box predicted by the $k$-th instance candidate and $N$ points in the input point cloud $\mathbf{P}$, or $f_{geo}^{(k,i)} = \left | f_{box}^{(i)}-f_{box}^{(k)} \right |$.



\subsection{Network Training}
\label{sec:loss}

We train our approach with the \textit{Pointwise Loss} and \textit{Instance-wise Loss}. 
The former is incurred at the point-wise prediction, i.e., the cross-entropy loss for semantic segmentation, and the L1 loss and gIoU loss \cite{Rezatofighi_2018_CVPR} for bounding box regression.
The latter is incurred at each instance prediction namely classification, box prediction, and mask prediction using the one-to-many matching loss proposed by \cite{jia2022detrs} for 2D object detection.
Specifically, the matching cost is the combination of instance classification and instance masks:
\begin{equation}
\label{eqn:matching_cost}
    C(k,j) = \gamma_{mask} C_{mask}(\widehat{m}^{(k)}, m^{(j)}) + C_{cls}(\widehat{l}^{(k)},l^{(j)}),
\end{equation}
where $C_{mask}$ is the dice loss \cite{sudre2017generalised} between two masks. Precisely, $S$ predicted masks are matched to one ground-truth mask by duplicating the ground truth $S$ times in Hungarian matching. In this way, the training convergence is much faster and the mask prediction performance is better than the one-to-one matching proposed by DETR \cite{carion2020end}.
Then the instance-wise loss incurred between the ground-truth masks and their matched predicted masks is defined as:
\begin{equation}
    \label{eqn:inst_loss}
    L_{inst} = L_{cls} + \lambda_{box}L_{box} + \lambda_{mask}L_{mask} + \lambda_{ms}L_{MS},
\end{equation}
where $L_{cls}$ is the cross-entropy loss, $L_{mask}$ is the combination of dice loss and BCE loss, $L_{box}$ is the combination of L1 loss and gIoU loss, and $L_{MS}$ is the Mask-Scoring loss \cite{huang2019msrcnn}.

\section{Experiments}
\label{sec:experiments}
\subsection{Experimental Setup}
\noindent\textbf{Datasets.} We evaluate our method on three datasets: ScanNetV2 \cite{dai2017scannet}, S3DIS \cite{armeni2017joint}, and STPLS3D \cite{chen2022stpls3d}. The \textit{ScanNetV2} dataset consists of 1201, 312, and 100 scans with 18 object classes for training, validation, and testing, respectively. 
We report the evaluation results on the validation and test sets of ScanNetV2 as in the previous work. 
The \textit{S3DIS} dataset contains 271 scenes from 6 areas with 13 categories. We report evaluations for both Area 5 and 6-fold cross-validation.
The \textit{STPLS3D} dataset is an aerial photogrammetry point cloud dataset from real-world and synthetic environments. It includes 25 urban scenes of a total of 6km$^2$ and 14 instance categories. Following \cite{chen2021hierarchical,vu2022softgroup}, we use scenes 5, 10, 15, 20, and 25 for validation and the rest for training.

\noindent\textbf{Evaluation Metrics.} Average precision commonly used for object detection and instance segmentation tasks is adopted, i.e., AP$_{50}$ and AP$_{25}$ are the scores with IoU thresholds of 50\% and 25\%, respectively, while AP is the averaged scores with IoU thresholds from 50\% to 95\% with a step size of 5\%. Box AP means the average precision of the 3D axis-aligned bounding box prediction. Additionally, the S3DIS is also evaluated using mean coverage (mCov), mean weighed coverage (mWCov), mean precision (mPrec$_{50}$), and mean recall (mRec$_{50}$) with IoU threshold of 50\%.

\noindent\textbf{Implementation Details.}
We implement our model using PyTorch deep learning framework \cite{paszke2017automatic} and train it on 320 epochs with AdamW optimizer on a single V100 GPU. The batch size is set to 16. The learning rate is initialized to 0.004 and scheduled by a cosine annealing \cite{zhang2021point}. Following \cite{vu2022softgroup}, we set the voxel size to 0.02m for ScanNetV2 and S3DIS, and to 0.3m for STPLS3D due to its sparsity and much larger scale. In training, the scenes are randomly cropped at a maximum number of 250,000 points. In testing, the whole scenes are fed into the network without cropping. 
We use the same backbone design as in \cite{vu2022softgroup}, which outputs a feature map of 32 channels. A stack of two layers of PA is used in the sampling-based instance-aware encoder. $\tau$ is set to 0.5. We set the ball query radius $r$ to 0.2 and 0.4 for these two layers and the number of neighbors $Q=32$ for both layers. 
We also implement the box-aware dynamic convolution with two layers with a hidden dimension of 32. $\gamma_{mask}$ is set to 5. $\lambda_{box}$, $\lambda_{mask}$, and $\lambda_{ms}$ are set to 1, 5, and 1, respectively.
In training, we set $S=4$ and $K=256$. In inference, we set $K=384$ and use Non-Maximum-Suppression to remove redundant mask predictions with a threshold of 0.2. Following \cite{han2020occuseg,liang2021instance,wu2022dknet}, we leverage the superpoints \cite{landrieu2018large,landrieu2019point} to align the final predicted masks on the ScanNetV2 dataset.

\subsection{Main Results}

\myheading{ScanNetV2.} We report the instance segmentation results on the hidden test set in Tab.~\ref{tab:scannet_test} and the instance segmentation and object detection results on the validation set in Tab.~\ref{tab:scannet_val}. On the hidden test set, \Approach~achieves 55.9/76.6 in AP/AP$_{50}$, set a new state-of-the-art performance on ScanNetV2 benchmark. On the validation set, our proposed method surpasses the second-best method with large margins, $+3.7/5.5/3.6$ in AP/AP$_{50}$/AP$_{25}$ and $+2.6/6.5$ in Box AP$_{50}$/Box AP$_{25}$.

\myheading{S3DIS.} Tab.~\ref{tab:s3dis} summarizes the results on Area 5 and 6-fold cross-validation of the S3DIS dataset. On both Area 5 and cross-validation evaluations, our proposed method overtakes the strongest method by large margins in almost metrics. On the 6-fold cross-validation evaluation, we achieve 74.9/76.8/77.1 in mCov/mWCov/mRec$_{50}$, with an improvement of 
$+3.5/2.7/3.1$ compared with the second-strongest method.

\myheading{STPLS3D.} Tab.~\ref{tab:stpls3d_val} shows the results on the validation set of the STPLS3D dataset. Our method achieves the highest performance in all metrics and surpasses the second-best by $+3.0/2.2$ in AP/AP$_{50}$.

\begin{table}
\small
\setlength{\tabcolsep}{8pt}
\centering
\begin{tabular}{lcccc}
\toprule
\textbf{Method} & \textbf{Venue} & \textbf{AP} & \textbf{AP$_{50}$} & \textbf{AP$_{25}$}  \\ 
\midrule
SGPN \cite{wang2018sgpn} & CVPR 18 & 4.9 & 14.3 & 26.1 \\
MTML \cite{yang2019learning} & ICCV 19 & 28.2 & 54.9 & 73.1 \\
3D-BoNet \cite{yang2019learning} & NeurIPS 19 & 25.3 & 48.8 & 68.7 \\
PointGroup \cite{jiang2020pointgroup} & CVPR 20 & 40.7 & 63.6 & 77.8 \\
OccuSeg \cite{han2020occuseg} & CVPR 20 & 44.3 & 67.2 & 74.2 \\
DyCo3D \cite{he2021dyco3d} & CVPR 21 & 39.5 & 64.1 & 76.1 \\
PE \cite{zhang2021point} & CVPR 21 & 39.6 & 64.5 & 77.6 \\
HAIS \cite{chen2021hierarchical} & ICCV 21 & 45.7 & 69.9 & 80.3 \\ 
SSTNet \cite{liang2021instance} & ICCV 21 & 50.6 & 69.8 & 78.9 \\
SoftGroup \cite{vu2022softgroup} & CVPR 22 & 50.4 & \underline{76.1} & \textbf{86.5} \\
RPGN \cite{dong2022rpgn} & ECCV 22 & 42.8 & 64.3 & 80.6 \\
PointInst3D \cite{He2022PointInst3DS3} & ECCV 22 & 43.8 & - & - \\
Di\&Co3D \cite{zhao2022divide} & ECCV 22 & 47.7 & 70.0 & 80.2 \\
DKNet \cite{wu2022dknet} & ECCV 22 & \underline{53.2} & 71.8 & 81.5 \\
\midrule
\textbf{\Approach} &  - & \textbf{55.9} & \textbf{76.3} & \underline{84.5} \\ 
\bottomrule
\end{tabular}
\vspace{-4pt}
\caption{3D instance segmentation results on ScanNetV2 hidden test set in terms of AP scores. The best results are in \textbf{bold} and the second best ones are in \underline{underlined}. Our proposed method achieves the highest AP, outperforming the previous strongest method.}
\label{tab:scannet_test}
\vspace{-16pt}
\end{table}

\begin{table}
\small
\setlength{\tabcolsep}{3.5pt}
\centering
\begin{tabular}{lccccc}
\toprule
\textbf{Method}   & \textbf{AP}     & \textbf{AP$_{50}$} & \textbf{AP$_{25}$} & \textbf{Box AP$_{50}$}     & \textbf{Box AP$_{25}$} \\ 
\midrule
GSPN \cite{yi2019gspn} & 19.3 & 37.8 & 53.4 & 10.8 & 19.8\\
PointGroup \cite{jiang2020pointgroup} & 34.8 & 51.7 & 71.3 & 48.9 & 61.5 \\
HAIS \cite{chen2021hierarchical} & 43.5 & 64.4 & 75.6 & 53.1 & 64.3 \\ 
DyCo3D \cite{he2021dyco3d} & 40.6 & 61.0 & - & 45.3 & 58.9\\
SSTNet \cite{liang2021instance} & 49.4 & 64.3 & 74.0 & 52.7 & 62.5 \\
SoftGroup \cite{vu2022softgroup} & 46.0 & \underline{67.6} & \underline{78.9} & \underline{59.4} & \underline{71.6}\\
RPGN \cite{dong2022rpgn} & - & 64.2 & - & - & -\\
PointInst3D \cite{He2022PointInst3DS3} & 45.6 & 63.7 & - & 51.0 & - \\
Di\&Co3D \cite{zhao2022divide} & 47.7 & 67.2 & 77.2 & - & - \\
DKNet \cite{wu2022dknet} & \underline{50.8} & 66.7 & 76.9 & 59.0 & 67.4 \\
\midrule
\textbf{\Approach} & \textbf{54.5} & \textbf{73.1}  & \textbf{82.5} & \textbf{62.0} & \textbf{78.1}\\
\bottomrule
\end{tabular}
\vspace{-4pt}
\caption{3D instance segmentation and 3D object detection results on ScanNetV2 validation set.}
\label{tab:scannet_val}
\vspace{-4pt}
\end{table}

\begin{table}
\small
\setlength{\tabcolsep}{1.4pt}
\centering
\begin{tabular}{lcccccc}
\toprule
\textbf{Method}  & \textbf{AP}     & \textbf{AP$_{50}$} & \textbf{mCov}    & \textbf{mWCov} & \textbf{mPrec$_{50}$} & \textbf{mRec$_{50}$} \\ 
\midrule
SGPN$^{\dagger}$ \cite{yi2019gspn} & - & - & 32.7 & 35.5 & 36.0 & 28.7\\
PointGroup$^{\dagger}$ \cite{jiang2020pointgroup} & - & 57.8 & - & - & 61.9 & 62.1 \\
HAIS$^{\dagger}$ \cite{chen2021hierarchical} & - & - & 64.3 & 66.0 & 71.1 & 65.0 \\ 
SSTNet$^{\dagger}$ \cite{liang2021instance} & 42.7 & 59.3 & - & - & 65.6 & 64.2 \\
SoftGroup$^{\dagger}$ \cite{vu2022softgroup} & 51.6 & \textbf{66.1} & \underline{66.1} & \underline{68.0} & \underline{73.6} & 66.6 \\
RPGN$^{\dagger}$ \cite{dong2022rpgn} & - & - & - & - & 64.0 & 63.0 \\
PointInst3D$^{\dagger}$ \cite{He2022PointInst3DS3} & - & - & 64.3 & 65.3 & 73.1 & 65.2 \\
Di\&Co3D$^{\dagger}$ \cite{zhao2022divide} & - & - & 65.5 & 66.1 & 63.9 & \underline{67.2} \\
DKNet$^{\dagger}$ \cite{wu2022dknet} & - & - & 64.7 & 65.6 & 70.8 & 65.3\\
\midrule
\textbf{\Approach}$^{\dagger}$         & \textbf{56.3} & \textbf{67.5} & \textbf{70.0} & \textbf{70.7} & 70.5 & \textbf{72.0} \\
\bottomrule
\toprule
SGPN$^{\ddagger}$ \cite{yi2019gspn} & - & 54.4 & 37.9 & 40.8 & 38.2 & 31.2 \\
3D-BoNet$^{\ddagger}$ \cite{yang2019learning} & - & - & - & - & 65.6 & 47.7  \\
PointGroup$^{\ddagger}$ \cite{jiang2020pointgroup} & - & 64.0 & - & - & 69.6 & 69.2 \\
OccuSeg$^{\ddagger}$ \cite{han2020occuseg} & - & - & - & - & 72.8 & 60.3\\
HAIS$^{\ddagger}$ \cite{chen2021hierarchical} & - & - & 67.0 & 70.4 & 73.2 & 69.4 \\ 
SSTNet$^{\ddagger}$ \cite{liang2021instance} & 54.1 & 67.8 & - & - & 73.5 & 73.4 \\
SoftGroup$^{\ddagger}$ \cite{vu2022softgroup} & 54.4 & 68.9 & 69.3 & 71.7 & 75.3 & 69.8 \\
RPGN$^{\ddagger}$ \cite{dong2022rpgn} & - & - & - & - & \textbf{84.5} & 70.5 \\
PointInst3D$^{\ddagger}$ \cite{He2022PointInst3DS3} & - & - & \underline{71.5} & \underline{74.1} & 76.4 & \underline{74.0} \\
DKNet$^{\ddagger}$ \cite{wu2022dknet} & - & - & 70.3 & 72.8 & 75.3 & 71.1 \\
\midrule
\textbf{\Approach}$^{\ddagger}$ & \textbf{60.8} & \textbf{70.5} & \textbf{74.9} & \textbf{76.8} & \underline{77.5} & \textbf{77.1} \\
\bottomrule
\end{tabular}
\vspace{-4pt}
\caption{3D instance segmentation results on S3DIS dataset. Methods marked with $^{\dagger}$ are evaluated on Area 5, and methods marked with $^{\ddagger}$ are evaluated on 6-fold cross-validation.}
\label{tab:s3dis}
\vspace{-7pt}
\end{table}

\begin{table}
    \parbox{.45\linewidth}{
        \small
        \setlength{\tabcolsep}{2pt}
        \centering
        \begin{tabular}{lcc}
        \toprule
                    & \textbf{AP}     & \textbf{AP$_{50}$} \\ 
        \midrule
        PointGroup\cite{jiang2020pointgroup} & 23.3 & 38.5 \\
        HAIS\cite{chen2021hierarchical}      & 35.1 & 46.7 \\ 
        SoftGroup\cite{vu2022softgroup}      & \underline{46.2} & \underline{61.8}  \\
        \midrule
        \textbf{\Approach} & \textbf{49.2} & \textbf{64.0} \\
        \bottomrule
        \end{tabular}      
        \caption{3D instance segmentation results on STPLS3D validation set.}
        \label{tab:stpls3d_val}
    }
    \hfill
    \parbox{.52\linewidth}{
        \small
        \setlength{\tabcolsep}{3pt}
        \centering
        \begin{tabular}{ccccc}
        \toprule
        BCE & Focal & Dice& \textbf{AP}     & \textbf{AP$_{50}$} \\ 
        \midrule
        \checkmark &  &  & 53.6 & 72.1 \\
         & \checkmark &  & 46.5 & 63.4 \\ 
            & \checkmark  & \checkmark  & 53.9 & 72.1 \\
        \midrule
        \checkmark & & \checkmark & \textbf{54.5} & \textbf{73.1} \\
        \bottomrule
        \end{tabular}
        \caption{Ablation study on different combinations of mask losses on ScanNetV2 validation set.}
        \label{tab:ablation_maskloss}
        }
\vspace{-18pt}
\end{table}


\subsection{Qualitative Results}
\begin{figure*}[t]
  \centering
  \includegraphics[width=0.92\linewidth]{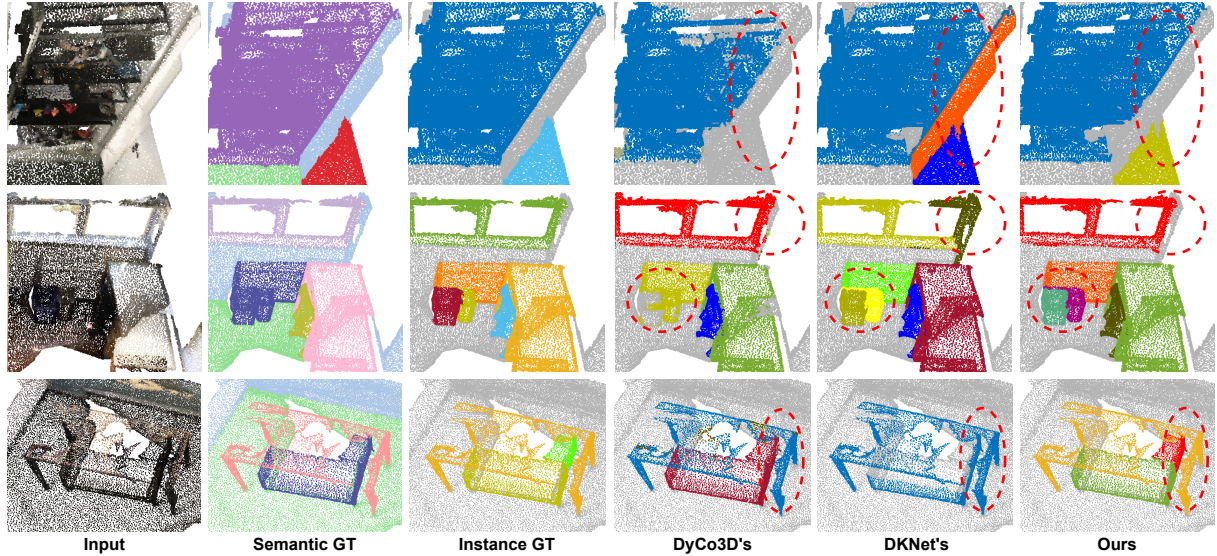}
\vspace{-8pt}
   \caption{Representative examples on ScanNetV2 validation set. Each row shows an example with the input, Semantic ground truth, and Instance ground truth in the first three columns. Our method (the last column) produces more precise instance masks, especially in regions where multiple instances with the same semantic label lie together.}
\vspace{-10pt}
   \label{fig:quali_scannetv2}
\end{figure*}

We visualize the qualitative results of our method, DyCo3D \cite{he2021dyco3d}, and DKNet \cite{wu2022dknet} on ScanNetV2 validation set in Fig.~\ref{fig:quali_scannetv2}. As can be seen, our method successfully distinguishes nearby instances with the same semantic class. Due to the limitation of clustering, DyCo3D \cite{he2021dyco3d} mis-segments parts of the bookshelf (row 1) and merges nearby sofas (rows 2, 3). DKNet \cite{wu2022dknet} over-segments the window in row 2, and also wrongly merges nearby sofas and table (row 3).


\subsection{Ablation Study}
We conduct a series of ablation studies on the validation set of the ScanNetV2 dataset to investigate \Approach.

\myheading{The impact of different combinations of mask losses} is shown in Tab.~\ref{tab:ablation_maskloss}. Notably, using a combination of binary cross entropy and dice loss yields the best result, 54.5 in AP.

\myheading{The impact of each component on the overall performance} is shown in Tab.~\ref{tab:ablation_components}. DyCo3D$^\ast$ in row 1 is our re-implementation of DyCo3D with the same backbone as \cite{chen2021hierarchical,vu2022softgroup,wu2022dknet}, and it is trained with the one-to-many matching loss.
The baseline in row 2 is a model with standard Farthest Point Sampling (FPS), standard Dynamic Convolution as in \cite{he2021dyco3d,He2022PointInst3DS3,wu2022dknet}, and without Local Aggregation Layer (LAL) 
. It can be seen that replacing the clustering and the tiny Unet in DyCo3D$^\ast$ decreases the performance from 49.4 to 47.9 in AP. When the standard FPS in the baseline is replaced by the Instance-aware Farthest Point Sampling (IA-FPS), the performance improves to 49.7 in row 3. When adding LAL to the baseline model, the AP score increases to 50.1 in row 4 and outperforms the AP of DyCo3D$^\ast$ by 0.7. 
Simply replacing the standard Dynamic Convolution with the Box-aware Dynamic Convolution (BA-DyCo) gains +0.7 in AP in row 5.
Especially, combining the IA-FPS and PA significantly boosts the performance, +5.5/+6.8 in AP/AP$_{50}$ in row 6. Finally, the full approach in row 7, \Approach~achieves the best performance 54.5/73.1 in AP/AP$_{50}$.


\begin{table}
\small
\setlength{\tabcolsep}{4.5pt}
\centering
\begin{tabular}{lcccccc}
\toprule
 & \textbf{IA-FPS} & \textbf{LAL} & \textbf{BA-DyCo} & \textbf{AP} & \textbf{AP$_{50}$} & \textbf{AP$_{25}$} \\
\midrule
DyCo3D$^\ast$ & & & &  49.4 & 67.6 & 77.4 \\
Baseline &  &  & & 47.9 & 66.4 & 77.1 \\
\midrule

 & \checkmark  &  &  & 49.7 & 67.5 & 78.6 \\
 &  & \checkmark & & 50.1 & 69.4 & 79.1 \\
 &  &  & \checkmark & 48.6 & 67.7 & 77.8 \\
 & \checkmark  & \checkmark &  & 53.4 & 71.9 & 81.8 \\
\midrule
\textbf{\Approach} & \checkmark & \checkmark  & \checkmark & \textbf{54.5} & \textbf{73.1} & \textbf{82.5} \\ 
\bottomrule
\end{tabular}
\vspace{-4pt}
\caption{Impact of each component of \Approach~on ScanNetV2 validation set. \textbf{IA-FPS}: Instance-aware Farthest Point Sampling, 
\textbf{LAL}: Local Aggregation Layer,
\textbf{BA-DyCo}: Box-aware Dynamic Convolution. $^\ast$: our improved version of DyCo3D \cite{he2021dyco3d}.}
\label{tab:ablation_components}
\vspace{-4pt}
\end{table}



\myheading{The impact of axis-aligned bounding box regression} is shown in Tab.~\ref{tab:ablation_box}. Without using the bounding box as an auxiliary supervision (\textbf{Aux.}), our method achieves 52.8/71.6 in AP/AP$_{50}$. Adding the bounding box loss during training brings a 0.6 improvement in AP. Especially, when using the bounding box as an extra geometric cue (\textbf{Geo. Cue}) in the dynamic convolution, the result significantly increases to 54.5/73.1 in AP/AP$_{50}$. This justifies our claim that the 3D bounding box is a critical geometric cue to distinguish instances in 3D point cloud.

\begin{table}[t]
    \parbox{.50\linewidth}{
        \small
        \setlength{\tabcolsep}{4pt}
        \centering
        \begin{tabular}{cccc}
        \toprule
        \textbf{Aux.} & \textbf{Geo. Cue} & \textbf{AP} & \textbf{AP$_{50}$} \\ 
        \midrule
            &   & 52.8  & 71.6 \\
        \checkmark  &   & 53.4  & 71.9 \\
        \checkmark  & \checkmark & \textbf{54.5} & \textbf{73.1} \\
        \bottomrule
        \end{tabular}
        \caption{The impact of 3D axis-aligned bounding box regression.}
        \label{tab:ablation_box}
    }
    \hfill
    \parbox{.44\linewidth}{
        \small
        \setlength{\tabcolsep}{7pt}
        \centering
        \begin{tabular}{ccc}
        \toprule
        \textbf{\# of PA} & \textbf{AP}     & \textbf{AP$_{50}$} \\ 
        \midrule
        1 & 53.2 & 72.5 \\
        2 & \textbf{54.5} & \textbf{73.1} \\
        3 & 54.3 & 73.0 \\ 
        \bottomrule
        \end{tabular}
        \caption{The number of Point Aggregator (\textbf{PA}) blocks.}
        \label{tab:ablation_pa}
        }
\vspace{-16pt}
\end{table}


\myheading{The impact of the number of Point Aggregator (PA) blocks} is represented in Tab.~\ref{tab:ablation_pa}. With a single block of PA, our method achieves 53.2/72.5 in AP/AP$_{50}$. Stacking two blocks of PA gives 1.3/0.6 gains in these metrics. However, when we add more blocks, the results slightly decrease to 54.3/73.0 in AP/AP$_{50}$.

\myheading{The impact of different designs of the Dynamic Convolution} is shown in Tab.~\ref{tab:ablation_dyco}. 
Here, using two layers of dynamic convolution with the hidden channels of 32 gives the best results. Using only a single layer of dynamic convolution leads to a significant drop in performance. On the other hand, adding too many layers, i.e., three layers yields worse results. Reducing the number of hidden channels slightly decreases the performance. Thanks to the additional geometric cue, even with only 216 parameters of dynamic convolution, our model can achieve 53.6/72.1 in AP/AP$_{50}$, demonstrating the robustness of the box-aware dynamic convolution.


\begin{table} 
\small
\centering
\setlength{\tabcolsep}{6pt}
\begin{tabular}{c|c|c|c|cc}
\toprule
\textbf{\# of layers} & \textbf{Dimensions} & \textbf{\# of params} & \textbf{AP}  & \textbf{AP$_{50}$} \\ \midrule
1 & (41,1) & \textbf{41}  & 45.7 & 67.1 \\ 
2 & (25,8,1) & 216 & 53.6 & 72.1 \\ 
2 & (41,16,1) & 688 & 53.9 & 72.3 \\ 
2 & (41,32,1) & 1376 & \textbf{54.5} & \textbf{73.1} \\ 
3 & (41,16,16,1) & 960 & 53.9 & 72.7 \\ 
3 & (41,32,16,1) & 1696 & 54.2 & 72.8 \\ 
\bottomrule
\end{tabular}
\vspace{-4pt}
\caption{Ablation on the Box-aware Dynamic Convolution.}
\label{tab:ablation_dyco}
\vspace{-4pt}
\end{table}

\myheading{Ablation on the chunk size of IA-FPS.} We study different designs of the sampling chunk size of IA-FPS in inference in Tab.~\ref{tab:ablation_iterative}. The first three rows show the results when we sample $K$ candidates at once. Increasing the number of samples from 256 to 384 slightly improves the overall performance, but at 512 samples, the result drops to 53.6 in AP. When splitting $K$ into smaller chunks of size (192,128,64) and sampling points based on Eq.~\eqref{eq:sampling}, the performance further boosts to 54.5/73.1 in AP/AP$_{50}$ in the last row.

\begin{table} 
\small
\centering
\setlength{\tabcolsep}{6pt}
\begin{tabular}{c|c|ccc}
\toprule
\textbf{Chunk size} & \textbf{Total samples $K$} &  \textbf{AP}  & \textbf{AP$_{50}$} & \textbf{AP$_{25}$} \\
\midrule
(256) & 256 & 53.9 & 72.2 & 80.8 \\ 
(384) & 384 & 54.2 & 72.4 & 81.4 \\ 
(512) & 512 & 53.6 & 71.9 & 81.1 \\ 
(128,128,128) & 384 & 54.0 & 72.8 & 81.0 \\
(192,128,64) & 384 & \textbf{54.5} & \textbf{73.1} &  \textbf{82.5} \\ 
\bottomrule
\end{tabular}
\vspace{-4pt}
\caption{Ablation on the sample chunk size of Iterative sampling.}
\label{tab:ablation_iterative}
\vspace{-16pt}
\end{table}

\begin{figure}[t]
  \centering
  \includegraphics[width=1.0\linewidth]{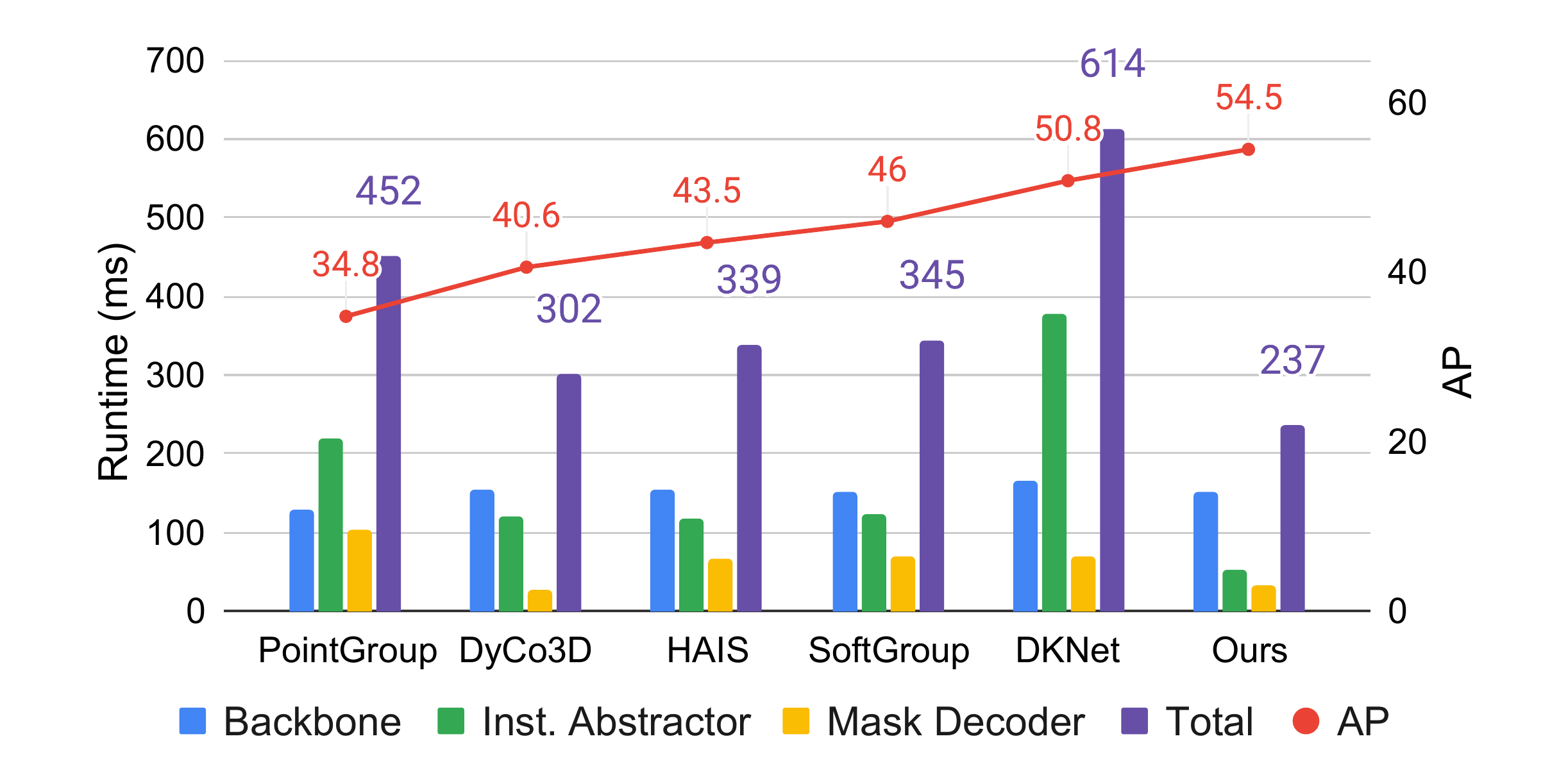}
  \vspace{-16pt}
    \caption{Components and total runtimes (in ms) and results in AP of five previous methods and \Approach~on ScanNetV2 validation set.}
   \label{fig:runtime_chart}
  \vspace{-6pt}
\end{figure}

\myheading{Runtime Analysis.} Fig.~\ref{fig:runtime_chart} reports the component and total runtimes of \Approach~and 5 recent state-of-the-art methods of 3DIS on the same Titan X GPU. All the methods can be roughly separated into three main stages: backbone, instance abstractor, and mask decoder. Our method is the fastest method, with only 237ms in total runtime and 152/53/32
ms in backbone/instance abstractor/mask decoder stages. Compared with the instance abstractors in PointGroup \cite{jiang2020pointgroup}, DyCo3D \cite{he2021dyco3d}, and SoftGroup \cite{vu2022softgroup} which are based on clustering, our instance abstractor based on our Point Aggregator significantly reduce the runtime. Our mask decoder, which is implemented by dynamic convolution, is the second fastest among these methods. This proves the efficiency of our proposed method. 

\section{Conclusion}
\label{sec:conclusion}
In this work, we have introduced the \Approach~, a concise dynamic convolution-based approach to address the task of 3D point cloud instance segmentation. Considering the performance of instance segmentation models relying on the recall of candidate queries, we propose our Instance-aware Farthest Point Sampling and Point Aggregator to efficiently sample candidates in the 3D point cloud. Additionally, leveraging the 3D bounding box as auxiliary supervision and a geometric cue for dynamic convolution further enhances the accuracy of our model. Extensive experiments on ScanNetV2, S3DIS, and STPLS3D datasets show that our approach achieves robust and significant performance gain on all datasets, surpassing state-of-the-art approaches in 3D instance segmentation by large margins, i.e., +2.7, +2.4, +3.0 in AP on ScanNetV2, S3DIS and STPLS3D. 

\begin{figure}[t]
  \centering
  \includegraphics[width=1.0\linewidth]{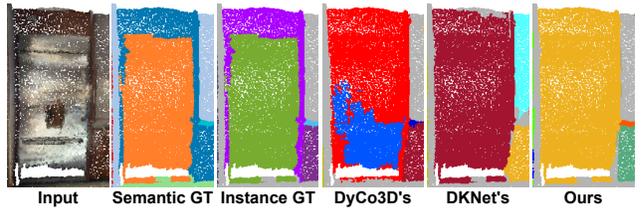}
\vspace{-18pt}
   \caption{A hard case on ScanNetV2 validation set where a fridge is bounded by a counter. \Approach~and previous methods wrongly merge points from these instances into a single object.}
\vspace{-14pt}
   \label{fig:quali_fail}
\end{figure}

Our method is not without limitations. For example, our instance-aware FPS does not guarantee to cover all instances as it relies on the current instance prediction to make decisions for point sampling. Our proposed axis-aligned bounding box may not tightly fit the shape of complicated instances. A hard case is shown in Fig.~\ref{fig:quali_fail} where a fridge is bounded by a counter. Our model cannot distinguish these points as they share similar bounding boxes. Addressing these limitations might lead to improvement in future work. Additionally, a new study on improving dynamic convolution by leveraging objects' geometric structures such as their shapes and sizes would be an interesting research topic.

{\small
\bibliographystyle{ieee_fullname}
\bibliography{egbib}
}
    
\newpage

\section{Supplementary Material}

In this supplementary material, we provide:
\begin{itemize}
    \setlength\itemsep{0em}
    \item The implementation details of the MLP in Point Aggregator and the late devoxelization (\cref{sec:implementation_detail}).
    \item Performances when using a smaller backbone (\cref{sec:small_backbone}).
    \item Per-class AP on the ScanNetV2 hidden set (\cref{sec:perclass_scannet}). 
    \item More qualitative results of our approach on  all test datasets (\cref{sec:more_qualitative}).
    \item The run-time analysis of various methods on the ScanNetV2 validation set (\cref{sec:runtime_analysis}).
\end{itemize}

\subsection{Implementation Details}

\label{sec:implementation_detail}

\myheading{The MLP in Point Aggregator (PA).}
\label{sec:mlp_lal}
The MLP in the PA consists of three blocks of Conv-BatchNorm-ReLU. The input channel to the MLP is $D+3$ while the hidden channel and the output channel are set to $D$.

\begin{figure}[t]
  \centering
  \includegraphics[width=0.8\linewidth]{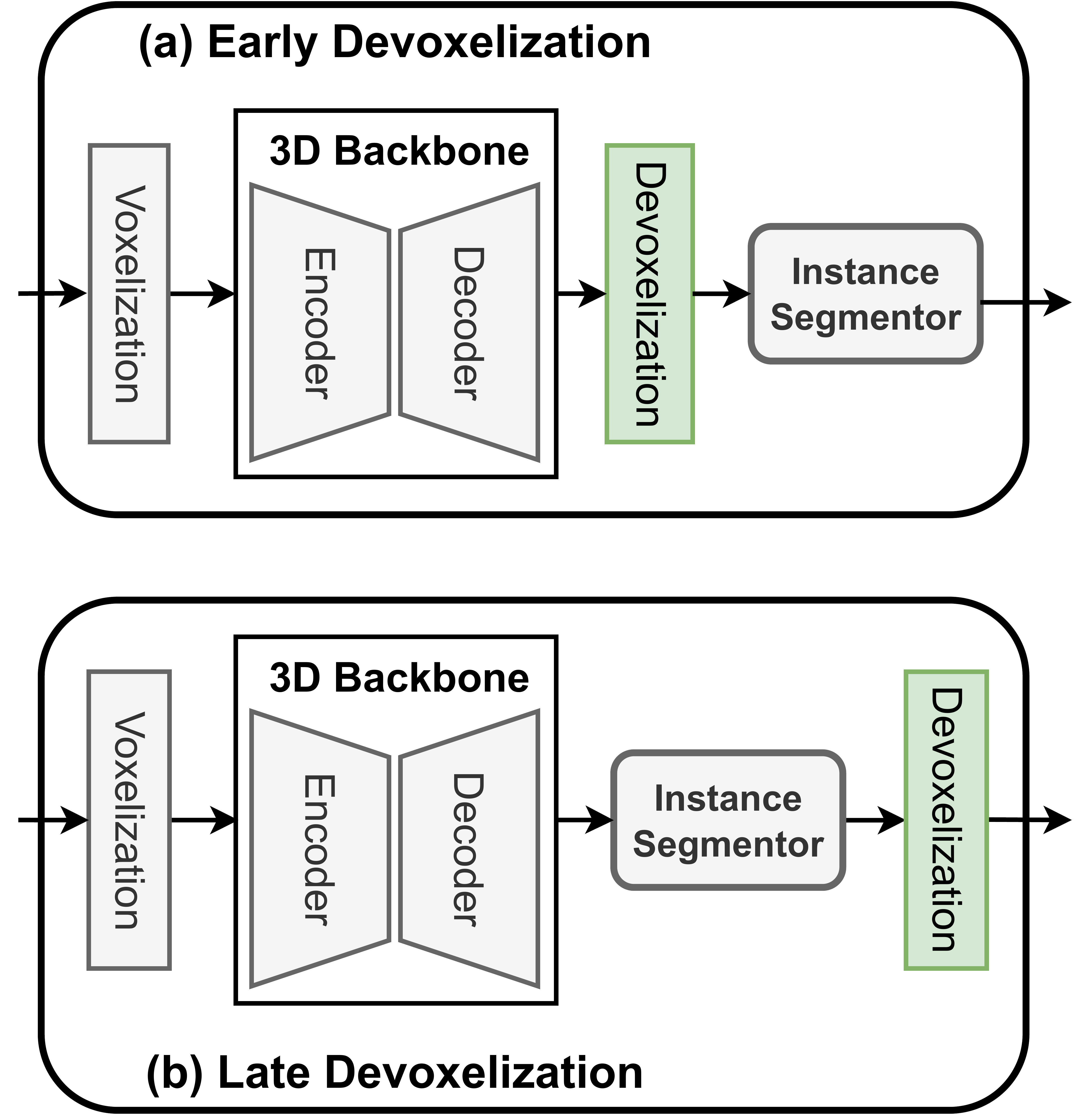}
   \caption{Difference between early and late devoxelization.}
   \label{fig:late_devoxel}
\end{figure}

\myheading{Late devoxelization.}
\label{sec:late_devoxel}
Recent methods \cite{chen2021hierarchical,vu2022softgroup,wu2022dknet} adopt the sparse convolutional network \cite{graham20183d} as their backbones. It requires the input point clouds to be voxelized as voxel grids and taken as input to the backbone network. The devoxelization step is performed right after the backbone to convert the voxel grids back to points. However, as all points in a voxel grid share the same features, the early devoxelization causes redundant computation and thus increases the memory consumption in later modules \cite{vu2022softgroup++}. Therefore, following \cite{vu2022softgroup++}, we employ the late devoxelization in our model in both training and testing. Fig.~\ref{fig:late_devoxel} illustrates the differences between late and early devoxelization.
The last two rows of Tab.~\ref{tab:runtime_detail} report the average inference time of our model using the early and late devoxelization, respectively. As can be seen, by late devoxelization, our model can reduce the run-time from 268 ms to 237 ms. We would note that applying the late devoxelization does not decrease the accuracy of our model. 

\subsection{Performances when using a smaller backbone}
\label{sec:small_backbone}
We report the performances of recent methods and \Approach using a smaller backbone as in DyCo3D on ScanNetV2 validation set in Tab~\ref{tab:abl_smallbb}. Our approach consistently outperforms others by a large margin on both AP/AP$_{50}$.

\subsection{Per-class AP on the ScanNetV2 Dataset} 
\label{sec:perclass_scannet}

We report the detailed results of the 18 classes on the ScanNetV2 hidden set in \cref{tab:scannet_test_detail}.

\subsection{More Qualitative Results of Our Approach}
\label{sec:more_qualitative} 

\begin{figure*}[h]
  \centering
  \includegraphics[width=1.0\linewidth]{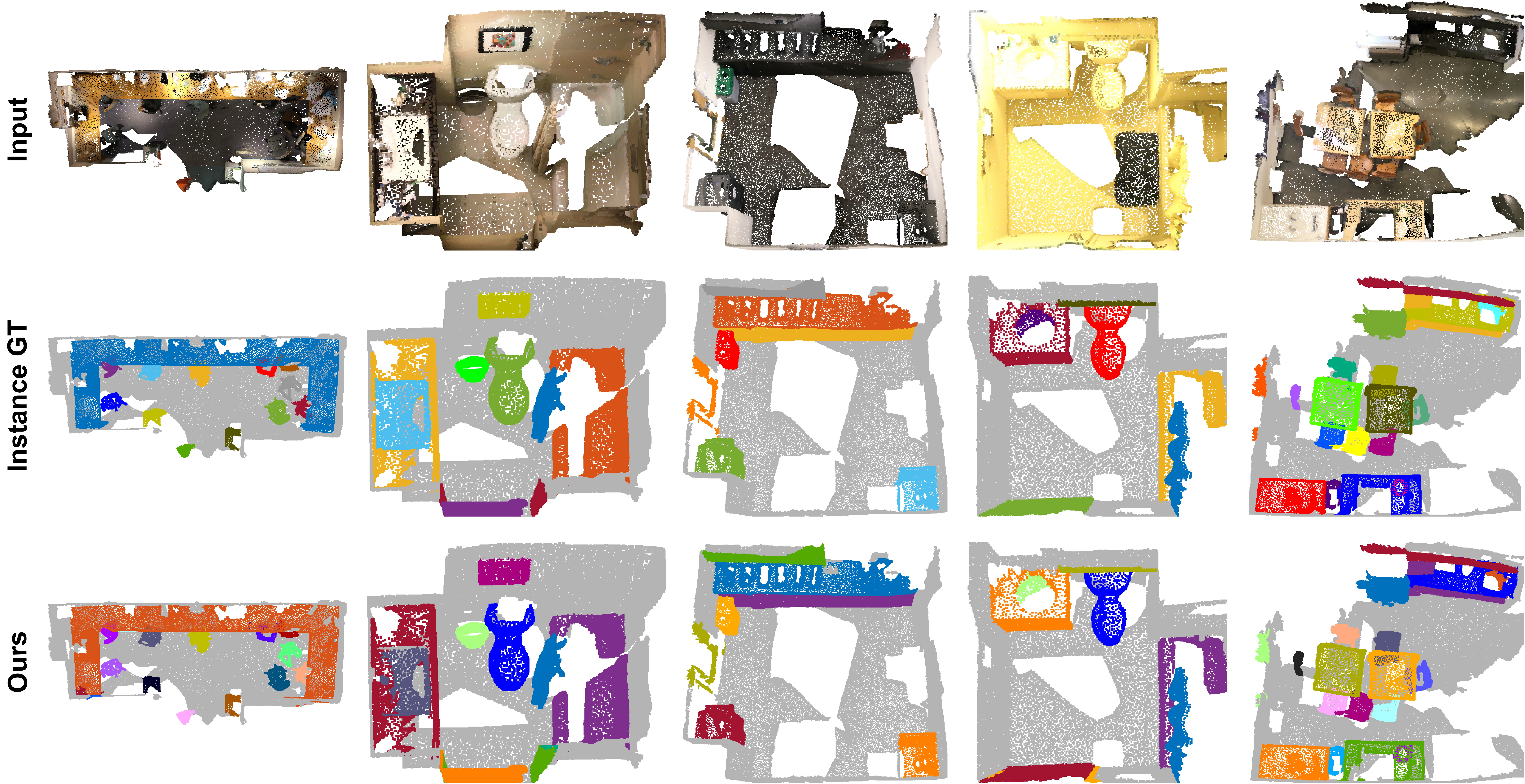}
   \caption{Qualitative results on ScanNetV2 dataset. Each column shows one example.}
   \label{fig:quali_scannet}
\end{figure*}

\begin{figure*}[h]
  \centering
  \includegraphics[width=1.0\linewidth]{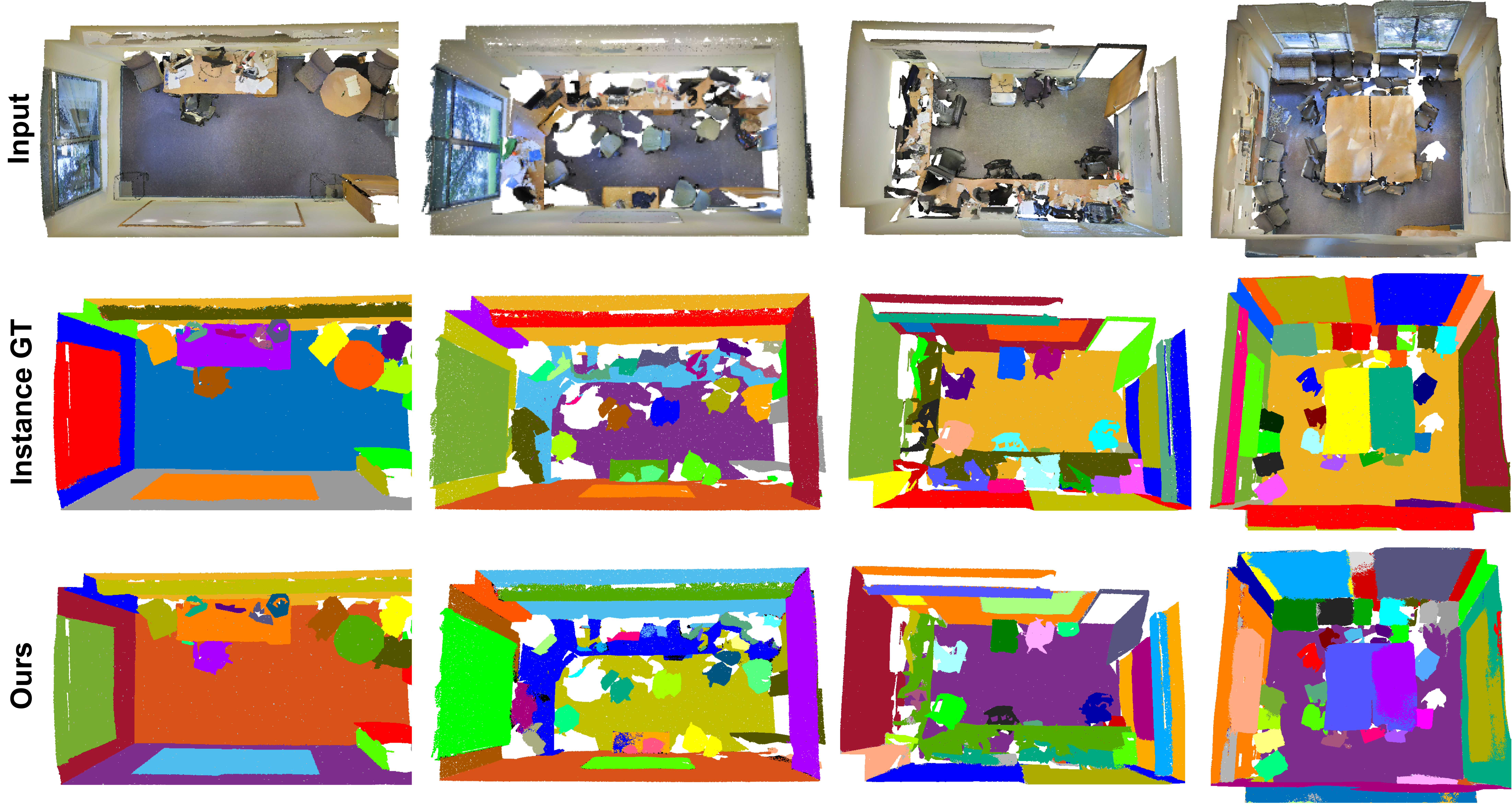}
   \caption{Qualitative results on S3DIS dataset. Each column shows one example.}
   \label{fig:quali_s3dis}
\end{figure*}

\begin{figure*}[h]
  \centering
  \includegraphics[width=0.95\linewidth]{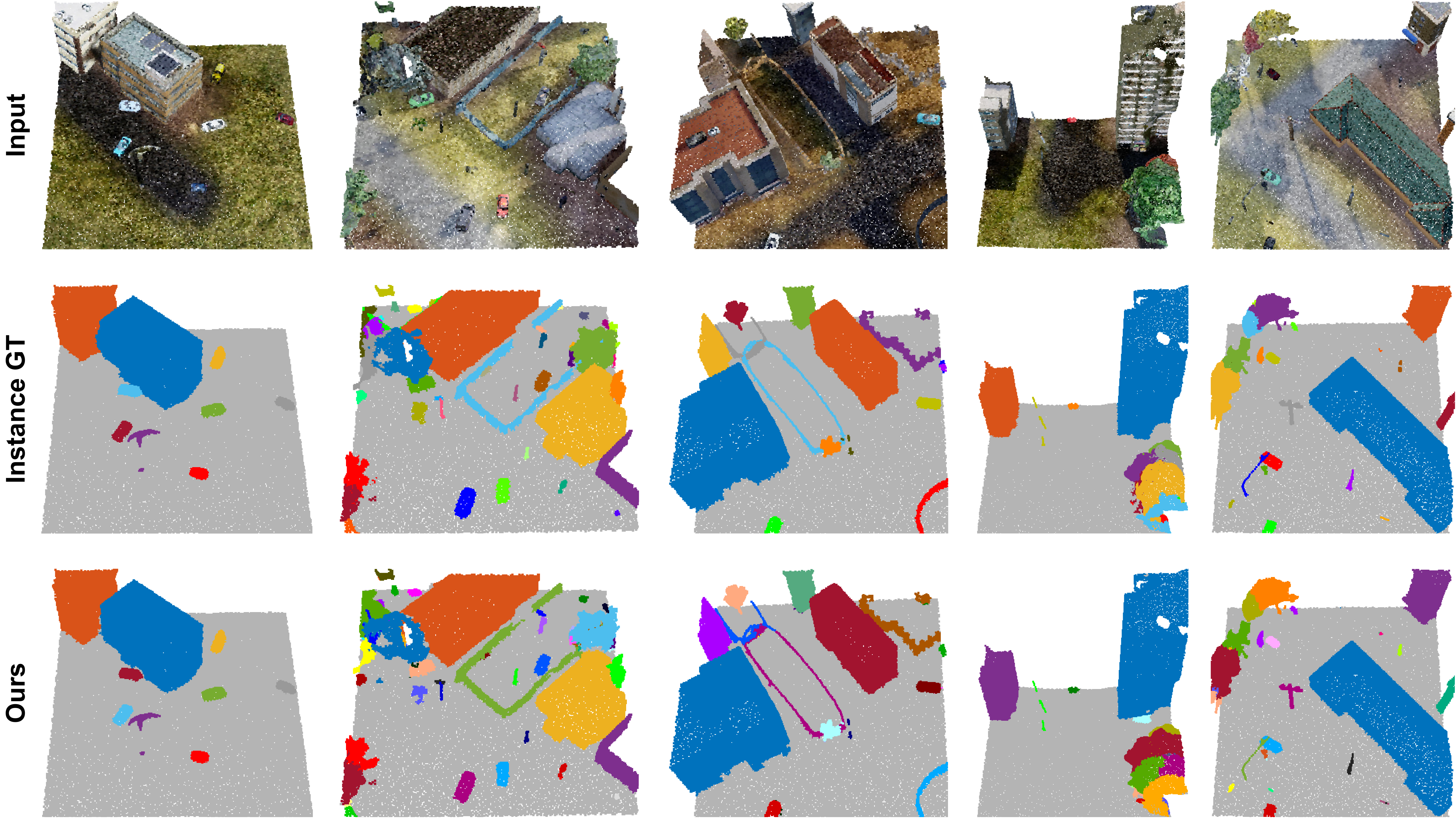}
   \caption{Qualitative results on STPLS3D dataset. Each column shows one example.}
   \label{fig:quali_stpls3d}
\end{figure*}

The qualitative results of our approach on the ScanNetV2, S3DIS, and STPLS3D datasets are visualized in \cref{fig:quali_scannet}, \cref{fig:quali_s3dis}, and \cref{fig:quali_stpls3d}, respectively.

\subsection{Run-Time Analysis} 
\label{sec:runtime_analysis}
\myheading{Training.} The training time for our model with the default setting on the ScanNetV2 \cite{dai2017scannet} training set is about 20 hours on a single NVIDIA V100 GPU.

\myheading{Inference.} \cref{tab:runtime_detail} shows the average inference time of each component and the whole approach for all scans of the ScanNetV2 validation set. The first 10 rows show the run-time analysis of 10 previous methods. Row 11 presents the run-time of our proposed method. The last row reports the run-time of our model without using late devoxelization (\cref{sec:late_devoxel}).

\begin{table}
    \small
    \centering
    \begin{tabular}{lcccc}
    \toprule
                & DyCo3D-S     & HAIS-S     & PointInst3D-S & \textbf{ISBNet-S} \\ 
    \midrule
    \textbf{AP} & 35.4 & 38.0 & 39.6 & \textbf{49.5} \\
    \textbf{AP$_{50}$} & 57.6 & 59.1   & 59.2 & \textbf{70.1}  \\
    \bottomrule
    \end{tabular}      
    \vspace{-4pt}
    \caption{3DIS results of recent methods with the same small backbone as used in DyCo3D on ScanNetV2 validation set.}
    \vspace{-17pt}
    \label{tab:abl_smallbb}
\end{table}

\begin{table*}[t]
\small
\setlength{\tabcolsep}{3.5pt}
\centering
\begin{tabular}{l|c|ccccccccccccccccccccc}
\toprule
Method & AP & \rotatebox{90}{bathtub} & \rotatebox{90}{bed} & \rotatebox{90}{bookshe.}      & \rotatebox{90}{cabinet}  & \rotatebox{90}{chair} & \rotatebox{90}{counter}  & \rotatebox{90}{curtain}  & \rotatebox{90}{desk}  & \rotatebox{90}{door}  & \rotatebox{90}{other}  & \rotatebox{90}{picture}  & \rotatebox{90}{fridge} & \rotatebox{90}{s.curtain}  & \rotatebox{90}{sink}  & \rotatebox{90}{sofa} & \rotatebox{90}{table} & \rotatebox{90}{toilet} & \rotatebox{90}{window} \\ 
\hline
SGPN \cite{wang2018sgpn} & 4.9 & 2.3 & 13.4 & 3.1 & 1.3 & 14.4 & 0.6 & 0.8 & 0.0 & 2.8 & 1.7 & 0.3 & 0.9 & 0.0 & 2.1 & 12.2 & 9.5 & 17.5 & 5.4 \\
3D-BoNet \cite{yang2019learning}& 25.3 & 51.9 & 32.4 & 25.1 & 13.7 & 34.5 & 3.1 & 41.9 & 6.9 & 16.2 & 13.1 & 5.2 & 20.2 & 33.8 & 14.7 & 30.1 & 30.3 & 65.1 & 17.8 \\
3D-MPA \cite{engelmann20203d}  & 35.5 & 45.7 & 48.4 & 29.9 & 27.7 & 59.1 & 4.7 & 33.2 & 21.2 & 21.7 & 27.8 & 19.3 & 41.3 & 41.0 & 19.5 & 57.4 & 35.2 & 84.9 & 21.3 \\
PointGroup \cite{jiang2020pointgroup}  & 40.7 & 63.9 & 49.6 & 41.5 & 24.3 & 64.5 & 2.1 & 57.0 & 11.4 & 21.1 & 35.9 & 21.7 & 42.8 & 66.0 & 25.6 & 56.2 & 34.1 & 86.0 & 29.1 \\
OccuSeg \cite{han2020occuseg} & 48.6 & 80.2 & 53.6 & 42.8 & 36.9 & 70.2 & \underline{20.5} & 33.1 & 30.1 & 37.9 & 47.4 & 32.7 & 43.7 & \textbf{86.2} & \underline{48.5} & 60.1 & 39.4 & 84.6 & 27.3 \\
DyCo3D \cite{he2021dyco3d}  & 39.5 & 64.2 & 51.8 & 44.7 & 25.9 & 66.6 & 5.0 & 25.1 & 16.6 & 23.1 & 36.2 & 23.2 & 33.1 & 53.5 & 22.9 & 58.7 & 43.8 & 85.0 & 31.7 \\
PE \cite{zhang2021point}  & 39.6 & 66.7 & 46.7 & 44.6 & 24.3 & 62.4 & 2.2 & 57.7 & 10.6 & 21.9 & 34.0 & 23.9 & 48.7 & 47.5 & 22.5 & 54.1 & 35.0 & 81.8 & 27.3 \\
HAIS \cite{chen2021hierarchical}  & 45.7 & 70.4 & 56.1 & 45.7 & 36.3 & 67.3 & 4.6 & 54.7 & 19.4 & 30.8 & 42.6 & 28.8 & 45.4 & 71.1 & 26.2 & 56.3 & 43.4 & 88.9 & 34.4 \\ 
SSTNet \cite{liang2021instance}  & 50.6 & 73.8 & 54.9 & \underline{49.7} & 31.6 & 69.3 & 17.8 & 37.7 & 19.8 & 33.0 & 46.3 & 57.6 & 51.5 & \underline{85.7} & \textbf{49.4} & 63.7 & 45.7 & 94.3 & 29.0 \\
SoftGroup \cite{vu2022softgroup} & 50.4 & 66.7 & 57.9 & 37.2 & \underline{38.1} & 69.4 & 7.2 & \textbf{67.7} & 30.3 & 38.7 & \textbf{53.1} & 31.9 & \underline{58.2} & 75.4 & 31.8 & \textbf{64.3} & 49.2 & 90.7 & \textbf{38.8} \\
RPGN \cite{dong2022rpgn}& 42.8 & 63.0 & 50.8 & 36.7 & 24.9 & 65.8 & 1.6 & \underline{67.3} & 13.1 & 23.4 & 38.3 & 27.0 & 43.4 & 74.8 & 27.4 & 60.9 & 40.6 & 84.2 & 26.7 \\
PointInst3D \cite{He2022PointInst3DS3}  & 43.8 & \underline{81.5} & 50.7 & 33.8 & 35.5 & 70.3 & 8.9 & 39.0 & 20.8 & 31.3 & 37.3 & 28.8 & 40.1 & 66.6 & 24.2 & 55.3 & 44.2 & \underline{91.3} & 29.3 \\
DKNet \cite{wu2022dknet}  & \underline{53.2} & \underline{81.5} & \textbf{62.4} & \textbf{51.7 }& 37.7 & \textbf{74.9} & 10.7 & 50.9 & \underline{30.4} & \underline{43.7} & 47.5 & \underline{58.1} & 53.9 & 77.5 & 33.9 & \underline{64.0} & \underline{50.6} & 90.1 & \underline{38.5} \\
\hline
\Approach & \textbf{55.9} & \textbf{92.6} & \underline{59.7} & 39.0 & \textbf{43.6} & \underline{72.2} & \textbf{27.6} & 55.6 & \textbf{38.0} & \textbf{45.0} & \underline{50.5} & \textbf{58.3} & \textbf{73.0} & 57.5 & 45.5 & 60.3 & \textbf{57.3} & \textbf{97.9} & 33.2 \\ 
\bottomrule
\end{tabular}
\caption{Per-class AP of 3D instance segmentation on the ScanNetV2 hidden test set. Our proposed method achieves the highest average AP and outperforms the previous strongest method significantly.}
\label{tab:scannet_test_detail}
\end{table*}

\begin{table}[t!]
\footnotesize
\setlength{\tabcolsep}{2pt}
\centering
\begin{tabular}{llc}
\toprule
\textbf{Method}  & \textbf{Component time} (ms) & \textbf{Total} (ms) \\ 
\midrule
\multirow{3}{*}{SGPN \cite{wang2018sgpn}} & Backbone (GPU): 2080 & \multirow{3}{*}{158439} \\
 &                      Group merging (CPU): 149000  &  \\
 &                      Block merging (CPU): 7119  &  \\
\midrule
\multirow{3}{*}{3D-BoNet \cite{hou20193d}} & Backbone (GPU): 2083 & \multirow{3}{*}{9202} \\
 &                      Group merging (CPU): 667  &  \\
 &                      Block merging (CPU): 7119  &  \\
\midrule
\multirow{3}{*}{OccuSeg \cite{han2020occuseg}} & Backbone (GPU): 189 & \multirow{3}{*}{1904} \\
 &                      Group merging (CPU): 1202  &  \\
 &                      Block merging (CPU): 513  &  \\
\midrule
\multirow{3}{*}{PointGroup \cite{jiang2020pointgroup}} & Backbone (GPU): 128 & \multirow{3}{*}{452} \\
 &                      Clustering (GPU+CPU): 221  &  \\
 &                      ScoreNet (GPU): 103  &  \\
\midrule
\multirow{3}{*}{SSTNet \cite{liang2021instance}} & Backbone (GPU): 125 & \multirow{3}{*}{428} \\
 &                      Tree net. (GPU+CPU): 229  &  \\
 &                      ScoreNet (GPU): 74  &  \\
\midrule
\multirow{3}{*}{HAIS \cite{chen2021hierarchical}} & Backbone (GPU): 154 & \multirow{3}{*}{339} \\
 &                      Hier. aggre. (GPU+CPU): 118  &  \\
 &                      ScoreNet (GPU): 67  &  \\
\midrule
\multirow{3}{*}{DyCo3D \cite{he2021dyco3d}} & Backbone (GPU): 154 & \multirow{3}{*}{302} \\
 &                      Weights Gen. (GPU+CPU): 120  &  \\
 &                      Dynamic Conv. (GPU): 28  &  \\
\midrule
\multirow{3}{*}{SoftGroup \cite{vu2022softgroup}} & Backbone (GPU): 152 & \multirow{3}{*}{345} \\
 &                      Soft grouping (GPU+CPU): 123  &  \\
 &                      Top-down refine. (GPU): 70  &  \\
\midrule
\multirow{3}{*}{Di\&Co \cite{zhao2022divide}} & Backbone (GPU): 163 & \multirow{3}{*}{502} \\
 &                      Group, Vote, Merge (GPU+CPU): 275  & \\
 &                      ScoreNet (GPU): 64  &  \\
\midrule
\multirow{3}{*}{DKNet \cite{wu2022dknet}} & Backbone (GPU): 165 & \multirow{3}{*}{614} \\
 &                      Cand. Mining \& Aggre. (GPU): 379  & \\
 &                      Dynamic Conv. + Postproc. (GPU): 70  &  \\
\midrule
\multirow{3}{*}{\textbf{ISBNet}} & Backbone (GPU): 152 & \multirow{3}{*}{\textbf{237}} \\
 &                      Point Pred. \& Inst. Enc. (GPU): 53  &  \\
 &                      Dynamic Conv. + Postproc. (GPU): 32  &  \\
\midrule
\multirow{3}{*}{\begin{minipage}{1in}\textbf{ISBNet} w/o \\ late voxelization\end{minipage}} & Backbone (GPU): 152 & \multirow{3}{*}{\textbf{268}} \\
 &                      Point Pred. \& Inst. Enc. (GPU): 82  &  \\
 &                      Dynamic Conv. + Postproc. (GPU): 34  &  \\
\bottomrule
\end{tabular}
\vspace{-0.1in}
\caption{Average inference time per scan of ScanNetV2 validation set on an NVIDIA Titan X GPU.}
\label{tab:runtime_detail}
\end{table}

\end{document}